\documentclass{article}
\usepackage{ijcai22}

\usepackage{times}

\usepackage{soul}
\usepackage{url}
\usepackage[hidelinks]{hyperref}
\usepackage[utf8]{inputenc}
\usepackage[small]{caption}
\usepackage{graphicx}
\usepackage{amsmath}
\usepackage{booktabs}
\usepackage{algorithm}
\usepackage{algorithmic}
\usepackage{multirow}
\usepackage{amsthm}
\usepackage{amssymb}
\usepackage{mathrsfs}
\usepackage{subfigure}

\title{\textsc{FedCat}: Towards Accurate Federated Learning via Device Concatenation}

\author{
Ming Hu\and
Tian Liu\and
Zhiwei Ling\and
Zhihao Yue\and
Mingsong Chen\footnote{Contact Author}\\
\affiliations
Shanghai Key   Lab   of   Trustworthy   Computing,   East   China   Normal   University,  Shanghai,  200062\\
\emails
mschen@sei.ecnu.edu.cn
}

\begin{document}

\maketitle

\begin{abstract}

As a promising distributed machine learning paradigm, 
Federated Learning (FL) 
enables all the involved
devices to train a global 
model collaboratively without exposing their local data privacy.
However, for non-IID scenarios, the classification 
accuracy of FL models decreases drastically due to the weight 
divergence caused by data heterogeneity.
Although various FL variants have been studied to 
improve model accuracy, most of them still suffer from the 
problem of non-negligible communication and computation overhead. In this paper, we introduce a novel FL approach named FedCat that can achieve high  model
accuracy based on
our proposed  device selection strategy  and device concatenation-based local training method. 
Unlike  conventional FL methods that aggregate local models trained on individual devices, FedCat  periodically aggregates local models after their traversals through a series of logically
concatenated devices, which can 
 effectively alleviate the model weight divergence problem. 
Comprehensive experimental results on four well-known benchmarks show that our approach can significantly improve the model accuracy of state-of-the-art  FL methods
without causing extra communication overhead. 

\end{abstract}

\section{Introduction}\label{sec:introduction}

Along with the prosperity of Artificial Intelligence (AI) 
techniques, 
Federated Learning (FL) is becoming a promising learning and  inference
paradigm in the  design of large-scale distributed AI applications~\cite{communication,hfl,spars}.
Unlike traditional centralized machine learning
methods, FL enables all the involved 
devices to train a global model collaboratively, while the training 
data is distributed on different local devices~\cite{practical,oneshot,fair}.
Rather than uploading
local data to a central cloud server, FL methods periodically
dispatch the global model from the cloud server to  devices for local training and then collect the locally
trained model to the cloud server for aggregation, thus the data privacy of devices can be guaranteed.

Although FL is promising in knowledge sharing among  devices, it 
greatly suffers from the problem of data heterogeneity~\cite{mimics,advance}.
Especially for 
non-Independent and Identically Distributed (non-IID)
scenarios, the model accuracy 
decreases drastically due to the notorious
``weight divergence'' phenomenon ~\cite{weight}.
For a non-IID scenario,
since different  devices  have different 
data distributions,  
the convergence directions of local models are inconsistent~\cite{convergence} 
and the aggregated global model may not fit for all the devices. 
Worse still, due to  data heterogeneity, 
it has been proved that the model 
convergence rate, in this case,  can be 
significantly influenced and slowed down~\cite{convergence}.

In order to improve the model performance in non-IID scenarios,  
various kinds of FL variants, e.g., device grouping-based methods~\cite{multicenter}, global control variable-based methods~\cite{scaffold,crosssilo}, and Knowledge Distillation (KD)-based methods~\cite{ensembledist,fedmd,datafree}, have been investigated 
to mitigate the data skew issue. 
Although these FL methods
can effectively promote
the classification accuracy, most of them inevitably
 introduce various side-effects, e.g., 
  extra communication and
 device-side computation overhead, and exposure of  local data privacy.
Such factors severely  make FL  unsuitable for safety-critical
  applications that  consist of  resource-constrained devices. {\it Therefore, how to design an FL method to achieve higher
classification accuracy without causing extra overhead for   devices is becoming a major bottleneck 
in the design of modern distributed
AI applications.}

To address the above challenge, we present a novel
FL method named FedCat that can significantly increase the 
classification accuracy  while alleviating the side-effects of data heterogeneity. Unlike traditional FL methods that train local 
models on individual devices, FedCat allows the local modeling training 
on a series of selected devices. By traversing through
the  datasets  of a series of 
concatenated devices, the  device models of FedCat can access much more data with less skewness for local training,
thus achieving
both high classification accuracy and 
fast training convergence. 
This paper makes the following three contributions:
\begin{itemize}
    \item 
    We present a novel FL architecture to facilitate the 
    local training on a series of concatenated devices in one training cycle, which can effectively mitigate the model weight divergence problem.  
    \item 
    To wisely mitigate the drawbacks of data heterogeneity, we propose a grouping and count-based device selection strategy, 
    which can substantially balance the distributions of training data 
    on concatenated
    devices while encouraging  all the  devices to sufficiently and 
    fairly participate  in the local training. 
    \item 
    We conduct both theoretical and empirical analysis on the  
     convergence rate of  FedCat, and  prove that FedCat converges as fast as FedAvg in arbitrarily heterogeneous data scenarios.
\end{itemize}
Experimental results on four well-known benchmarks
show that, compared with both the vanilla
FL (FedAvg) and state-of-the-art FL methods, FedCat can fastly 
achieve better model performance without incurring  extra communication and device-side computation overhead.

The rest of this paper is organized as follows. 
After the introduction to related work in Section~\ref{sec:relatedwork}, 
Section~\ref{sec:approach} details  the implementation 
of \textsc{FedCat} approach. 
Section~\ref{sec:experiment} shows the experimental results.
Finally, Section~\ref{sec:conclusions} concludes the paper.
\section{Related Work}\label{sec:relatedwork}

Although FL has the advantages of lower communication overhead and better data privacy protection than conventional distributed machine learning methods, it still suffers from the problem of
low classification accuracy due to heterogeneous  data 
on devices.
To address this problem, various methods have
been investigated, which can be classified into three categories, i.e., KD-based methods, global control variable-based methods, and device grouping-based methods.

The KD-based methods adopt  soft targets generated by the ``teacher model'' to guide the training of  ``student models''. For example, by leveraging a proxy dataset, Zhu et al.
[\citeyear{datafree}] proposed a data-free knowledge
distillation method named FedGen to address the heterogeneous FL problem using a built-in generator.
With ensemble distillation, FedDF~\cite{ensembledist} trained the central model through unlabeled data on the outputs of local models, which can significantly accelerate the model training.
Based on transfer learning and knowledge distillation, FedMD~\cite{fedmd} trained models on both public datasets and private datasets to mitigate the data heterogeneity.
However, all these  methods are required
to upload/dispatch generators or build public datasets for model training, which introduce significant
communication overhead and the risk of data privacy exposure.

The global control variable-based methods usually need to modify the penalty term of the loss function during the model training process.
As an example, FedProx~\cite{fedprox} regularized local loss functions with the square distance between local models and the global model, which stabilizes the model convergence using a proximal term.
By using global control variables, 
Karimireddyet al. [\citeyear{scaffold}] proposed a method named SCAFFOLD to correct the ``client-drift'' in the local training process. Wang et al. 
[\citeyear{imbalance}] presented a novel architecture to infer the composition of training data for each FL round, which can alleviate the impact of the imbalance issue.
However, all the methods above may result in non-negligible communication and computation overhead, while the improvements of these methods are limited.

The device grouping-based methods take the data similarity between all the devices into account.
For example, by grouping devices into multiple clusters, each learning round of FedCluster~\cite{fedcluster} consisted of multiple cycles of meta-update that boost the overall convergence.
Fraboni et al. [\cite{clustered}] introduced a client clustering method based on either sample size or model similarity, which leads to better client representativity and a reduced variance of client stochastic aggregation weights in FL.
However, the limitation of  communication resources  make the
above methods infeasible for large-scale FL systems.

To the best of our knowledge, our work is the first attempt that combines the merits of both device grouping and  concatenation to mitigate the weight divergence problem in local training
caused by data heterogeneity.
Due to the large size of training data
formed by the concatenated devices, 
our FedCat approach can quickly train a global model that has much better prediction performance than  the ones of
state-of-the-art methods 
 without causing extra  communication and  device-side computation overhead.

\section{Our FedCat Approach}\label{sec:approach}

Unlike conventional FL methods that aggregate  
local models trained on individual devices, 
 FedCat periodically aggregates local models after their traversal
 through a series of concatenated
 devices, where the local training is conducted 
 device by device along with the traversal.
Similar to  conventional FL methods, the global and local optimization objectives of FedCat are defined as follows:
\begin{footnotesize}
\begin{equation}
\vspace{-0.2in}
\begin{split}
\min_{w} F(w) = \sum_{i = 1}^{N} p_i f_i(w), \ s.t., \ p_i = \frac{n_i}{\sum_{i=1}^{N} n_i}
\end{split}
\end{equation}
\end{footnotesize}
\begin{footnotesize}
\begin{equation}
\begin{split}
f_i(w) = \frac{1}{n_i} \sum_{j = 1}^{n_i} \ell (w;x_j;y_j)
\end{split}
\end{equation}
\end{footnotesize}
where $N$ is the total number of devices, $p_i$ is the weight of the $i^{th}$ device that depends on the number of samples
(i.e., $n_i$) in the $i^{th}$ device, $\ell$ denotes a kind of customer-defined loss function (e.g. cross-entropy loss).
\begin{figure}[h]
	\vspace{-0.1in}
	\begin{center}
		\includegraphics[width=0.45\textwidth]{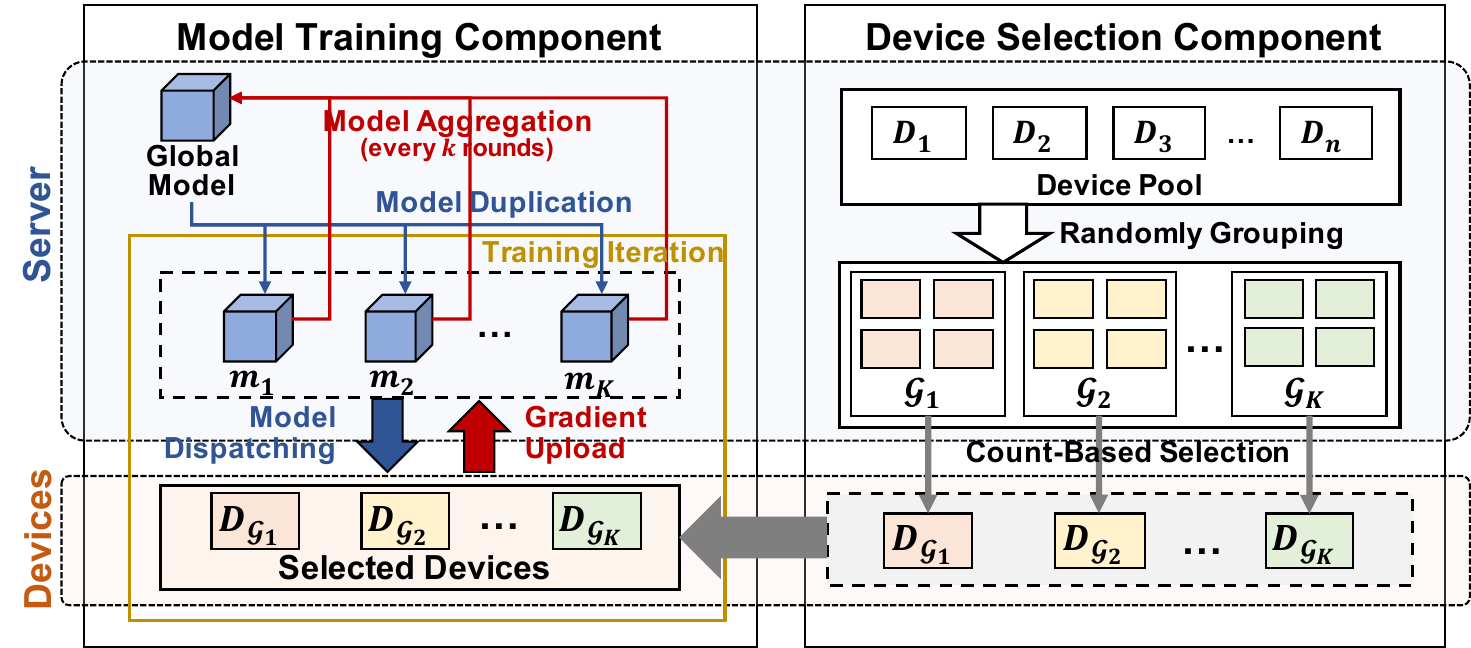}
			\vspace{-0.1in}
		\caption{Architecture and Workflow of  FedCat Framework} \label{fig:framework}
	\end{center}
	\vspace{-0.25in}
\end{figure}

\begin{figure*}[h]
	\begin{center}
		\includegraphics[width=5.3 in]{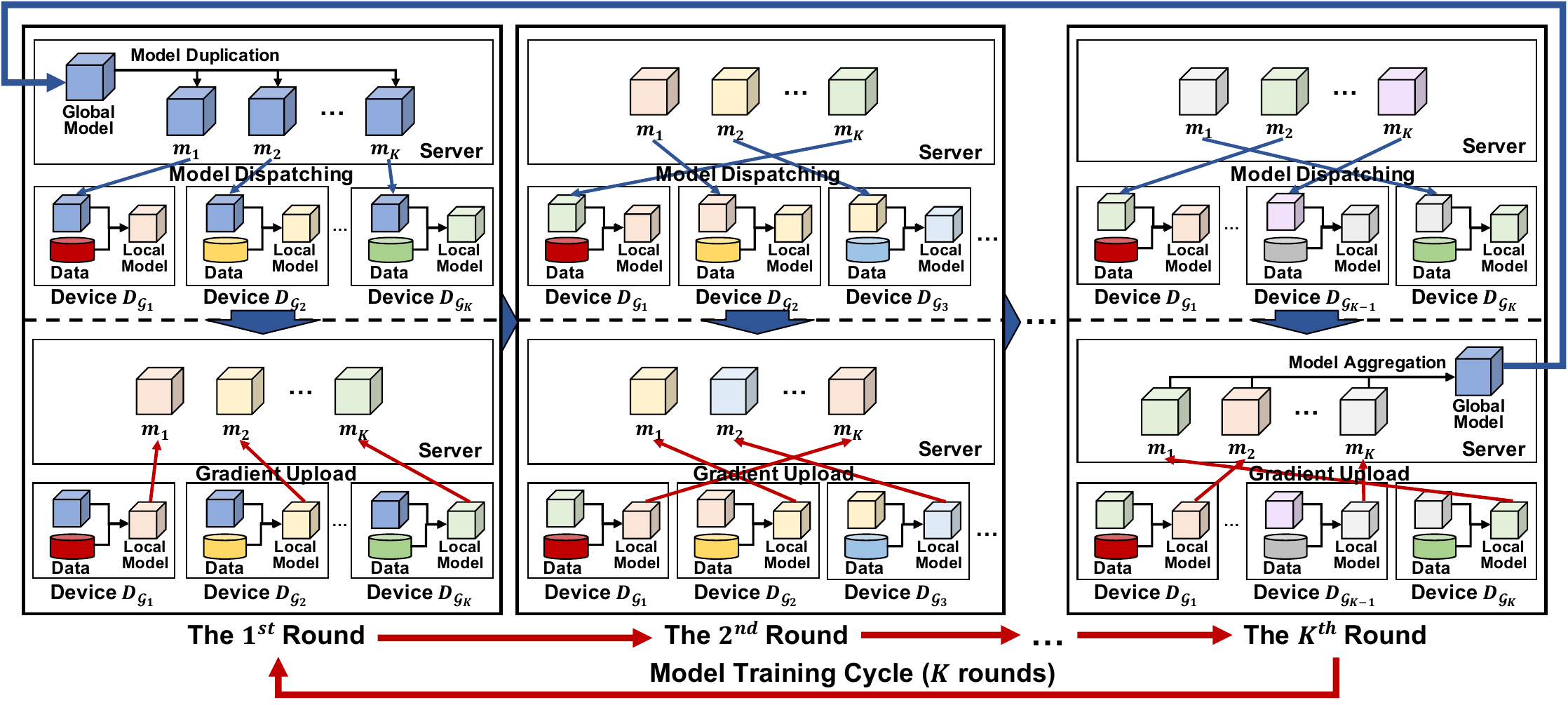}
			\vspace{-0.1in}
		\caption{Workflow of Device Concatenation-based Model Training} \label{fig:concat}
	\end{center}
		\vspace{-0.25in}
\end{figure*}

 \subsection{FedCat Implementation}
 
Figure \ref{fig:framework} shows an
overview of the architecture and workflow of our FedCat approach. 
The framework consists of two components, i.e., the device selection component and the model training component.
At the beginning of each training round, the device selection component randomly groups all the devices and conducts device selection
from these groups.
The model training component performs the local training on selected devices, where each model training cycle includes multiple training rounds.
Here, a model training cycle consists of three stages, i.e.,
model duplication, training iteration, and model aggregation.
The model duplication stage generates $K$ copies of the global model.
The training iteration stage dispatches the copies to the selected devices for local training and updates the copies after gradient collection.
At the end of each cycle, the server obtains a new global model by aggregating all the latest model copies.

\begin{algorithm}[htb]
\caption{Workflow of FedCat}
\label{alg:workflow}
\textbf{Input}: i) $rnd$, the total number of training rounds; ii) $S_{dev}$, the set of devices; iii) $K$, the number of participating devices in a training round\\
\textbf{Output}: i) $m_{glb}$, the trained global model\\
\textbf{FedCat}($rnd$,$S_{dev}$,$K$)
\begin{algorithmic}[1] 
\STATE $r \leftarrow 0$, $N \leftarrow len(S_{dev})$\;
\STATE $M_{dev}\leftarrow Map[N][K]$, $L_m \leftarrow []$\;
\STATE Initialize model $m_{glb}$\;
\FOR{$r = 0, 1, ..., rnd-1$}
\IF{$r\%(\lambda\times K)=0$}\label{line:devSelectStart}\label{line:devGroupStart}\label{line:roundStart}
\STATE Randomly divide $S_{dev}$ into $\mathcal{G}=\{\mathcal{G}_1,...,\mathcal{G}_K$\}\;
\ENDIF\label{line:devGroupEnd}
\STATE $L_{dev},M_{dev}\leftarrow DevSelect(\mathcal{G},M_{dev},r,K)$\label{line:devSelectEnd}\;
\IF {$r \% K = 0$}\label{line:trainStart}
\STATE Duplicate $K$ copies of $m_{glb}$ in $L_m$\label{line:modelDup}\;
\ENDIF
\STATE {\it offset}$ \leftarrow r \% K$\;
\STATE $L_m \leftarrow TrainIter(L_{dev},L_m,${\it offset}$)$\label{line:devTrain}\;
\IF {$r \% K = K-1$}
\STATE $m_{glb}\leftarrow Aggregation(L_m)$\label{line:modelAggr}\;
\ENDIF\label{line:trainEnd}
\STATE $r\leftarrow r+1$\label{line:roundEnd}\;
\ENDFOR
\STATE \textbf{return} $m_{glb}$\;
\end{algorithmic}
\end{algorithm}

Algorithm \ref{alg:workflow} presents the workflow of FedCat in detail.
Lines \ref{line:roundStart}-\ref{line:roundEnd} show the 
details of one FL round 
for FedCat including both 
 device selection and model training processes.
Lines \ref{line:devSelectStart}-\ref{line:devSelectEnd} present the device selection process in each training round, where we randomly divide all the devices into $K$ groups and select one device from each group for local training. 
The function $DevSelect$ denotes the device selection strategy which is implemented in Algorithm \ref{alg:devSel}.
$L_{dev}$ is the list of selected devices and $M_{dev}$ is a map that records the participation per device.
Note that we regroup devices in every $\lambda\times k$ rounds, where $\lambda$ is a hyperparameter that controls the frequency of device grouping.
Lines \ref{line:trainStart}-\ref{line:trainEnd} denote the model training process. Line \ref{line:modelDup} generates $K$ copies of the global model $m_{glb}$ in every $K$ rounds.
In Line \ref{line:devTrain}, $L_m$ is a list of 2-tuples in the form of $\langle m,d\rangle$, where $m$ indicates a model with $d$ accumulated training data.
The function $TrainIter$ denotes one training iteration stage, which dispatches models in $L_m$ to selected devices in $L_{dev}$ and updates $L_m$ after collecting the trained local models.
The details of $TrainIter$ are shown in Algorithm \ref{alg:devCat}.
Note that we use a variable (i.e., {\it offset})  of integer type to guide the model dispatching and gradient collection
processes.
Line \ref{line:modelAggr} denotes the model aggregation stage
that
aggregates models according to the size of accumulated training data.
We aggregate models in every $K$ rounds and
the $Aggregation$ function is defined as
\begin{align}
    Aggregation(L_m)=& \frac{\sum_{i=1}^K d_i\cdot m_i}{\sum_{j=1}^K d_j},
\end{align}
where $m_i$ indicates the $i^{th}$ model in $L_m$
and $d_i$ denotes  
the size of the training data  set used by $m_i$.

\subsection{Device Concatenation-Based Model Training}\label{sec:concat}
Figure \ref{fig:concat} shows the workflow of our device concatenation-based model training strategy. Each model training cycle contains $K$ model training rounds, where $K$ equals the number of selected devices in each round.
The server duplicates $k$ copies  of the global model in the first round of each model training cycle. 
After that, the server conducts $K$ training iterations, where each training iteration involves three steps:
\begin{itemize}
\item {\bf Step 1}: The server dispatches all the model copies to selected devices. In our approach, we dispatch the $i^{th}$ model copy to the $((i+j-2)\%K+1)^{th}$ selected device in the $j^{th}$ round of a model training cycle. 
For example, in Figure \ref{fig:concat}, $m_1$ is dispatched to the second device $D_{\mathcal{G}_2}$ in the second round.
\item {\bf Step 2}: The selected devices train received models using their local data and upload their gradients after local training finish.
\item {\bf Step 3}: The server updates all the model copies with the corresponding gradients. As an example shown
 in Figure \ref{fig:concat}, in the second round, the server updates $m_1$ using the gradient uploaded by the second device $D_{\mathcal{G}_2}$.
\end{itemize}
Note that, in the last round (i.e., the $K^{th}$ round) of each model training cycle, the aggregation stage aggregates
all the updated model copies.


\begin{algorithm}[htb]
\caption{Training Iteration}
\label{alg:devCat}
\textbf{Input}: i) $L_{dev}$, the list of selected devices; ii) $L_{m}$, the initial list of models with the sizes of their accumulated training data; iii) {\it offset},    device index offset for model dispatching\\
\textbf{Output}: $L_{m}$, an updated list of models with the sizes of their accumulated training data\\
\textbf{TrainIter}($L_{dev}$,$L_{m}$,{\it offset})
\begin{algorithmic}[1] 
\STATE $K\leftarrow len(L_m)$\;
\FOR{$\langle m_i,d_i \rangle$ in $L_m$}\label{line:dispatchStart}
\STATE $tag \leftarrow (${\it offset}$ + i)\%K$\;
\STATE $dev\leftarrow L_{dev}[tag]$\;
\STATE $Dispatch(tag,m_i,dev)$\;
\ENDFOR\label{line:dispatchEnd}
\STATE $L_{\nabla},L_d\leftarrow GradientCollect()$\label{line:gather}
\FOR{each $\langle m_i, d_i \rangle$ in $L_m$} \label{line:updateStart}
\STATE $tag \leftarrow (${\it offset}$ + i)\%K$\;
\STATE $m_i \leftarrow m_i+L_{\nabla}[tag]$\; \label{line:updateModel}
\STATE $d_i \leftarrow d_i+L_d[tag]$ \;\label{line:updateData}
\ENDFOR\label{line:updateEnd}
\STATE \textbf{return} $L_{m}$\;
\end{algorithmic}
\end{algorithm}

Algorithm \ref{alg:devCat} shows the details of a training iteration. Lines \ref{line:dispatchStart}-\ref{line:dispatchEnd} denote the model dispatching process, where we use {\it offset} to guide the model dispatching.
The function $GradientCollect$ in Line \ref{line:gather} collects the uploaded gradients and the sizes of training data from selected devices.
The list $L_{\nabla}$ stores the uploaded gradients and the list $L_d$ stores the sizes of local training datasets.
Lines \ref{line:updateStart}-\ref{line:updateEnd} denote the model updating process.
Line \ref{line:updateModel} updates the $i^{th}$ model copy $m_i$ 
using the corresponding uploaded gradient.
Line \ref{line:updateData} updates the size of accumulated training data  by adding the size of training data to it.

\subsection{Grouping- and Count-Based Device Selection}
To encourage more devices to participate in model training and avoid training one model with the same device multiple times in one training cycle, we propose a grouping- and count-based device selection strategy. 
We define a count table $M_{dev}$ to count the device participation in each round of the training cycle. Since each training cycle contains $K$ rounds, the count table is a matrix of size $N\times K$ where $N$ is the number of devices.
We use MBIE-EB \cite{count} to measure the weight of candidate devices, which is defined as:
\begin{equation}
w_{dev} = \frac{1}{\sqrt{M_{dev}[dev][{\it offset}]}}.
\end{equation}
We use $\epsilon$-greedy policy to select a device. It means that we select the device with the maximum weight having a probability of $\epsilon$ and select the device according to the following probability distribution with a probability of $1-\epsilon$:
\begin{equation}\label{equa:weightP}
P(dev) = \frac{w_{dev}}{\sum_{dev \in \mathcal{G}_i} w_{dev}}.
\end{equation}

Algorithm \ref{alg:devSel} details
our grouping- and count-based device selection strategy. $\mathcal{G}$ is the set of device groups. $L_{dev}$ is a list to store selected devices. 
The table $M_{dev}$ is our defined count table. 
Lines \ref{line:devSelStart}-\ref{line:devSelEnd} show the device selection details. In Line \ref{line:devSelWeight}, we select a device from group $\mathcal{G}_i$ according to the  policy $\pi_\theta$ based on the distribution defined in equation \ref{equa:weightP}. 
Line \ref{line:devSelGreedy} selects the device with the maximum weight from group $\mathcal{G}_i$.
Line \ref{line:updateMap} updates count table after device selection.

\begin{algorithm}[htb]
\caption{Device Selection Strategy}
\label{alg:devSel}
\textbf{Input}: i) $rnd$, the number of FL training round; ii) $S_{dev}$, the set of devices; iii) $K$, the number of devices which participate training in a round; iv) $\mathcal{G}$, the set of device groups\\
\textbf{Output}: i) $L_{dev}$, a list of selected devices; ii) $M_{dev}$, an updated count map\\
\textbf{DevSelect}($S_{dev}$,$M_{dev}$,$r$,$K$)
\begin{algorithmic}[1] 
\STATE $L_{dev}\leftarrow []$\;
\FOR{each group $\mathcal{G}_i$ in $\mathcal{G}=\{\mathcal{G}_1,...\mathcal{G}_K$\}}\label{line:devSelStart}
\IF{$random(0,1.0)>\epsilon$}
\STATE select $dev\sim \pi_\theta(\mathcal{G}_i,M_{dev})$\;\label{line:devSelWeight}
\ELSE
\STATE $dev\leftarrow \arg\max\limits_{dev\in \mathcal{G}_i}{w_{dev}}$ \;\label{line:devSelGreedy}
\ENDIF
\STATE $L_{dev}.append(dev)$\;
\STATE $M_{dev}[dev][r\%K]\leftarrow M_{dev}[dev][r\%K]+1$\;\label{line:updateMap}
\ENDFOR\label{line:devSelEnd}
\STATE \textbf{return} $L_{dev}$, $M_{dev}$\;
\end{algorithmic}
\end{algorithm}

\begin{table*}[t]
\centering
\footnotesize
\begin{tabular}{|c|c|c|c|c|c|c|c|}
\hline
\multirow{2}{*}{Dataset} & Heterogeneity & \multicolumn{6}{c|}{Test Accuracy (\%)} \\
\cline{3-8}
 & Settings & FedAvg & FedProx &  SCAFFOLD & FedGen & CluSamp & FedCat (Ours)\\
\hline
\hline
\multirow{3}{*}{MNIST} & $\alpha=0.1$      & $99.04$    & $99.02$     & $98.92$       & $99.18$  & $98.98$  & ${\bf 99.21}$\\
                       & $\alpha=0.5$      & $99.31$    & $99.27$     & $99.28$       & $99.41$ & $99.27$ & ${\bf 99.42}$\\
                       & $\alpha=1.0$      & $99.37$    & $99.30$     & $99.32$       & $99.37$  & $99.33$  & ${\bf 99.4}$\\
\hline
\multirow{3}{*}{CIFAR-10} & $\alpha=0.1$& $51.01$& $51.85$& $52.81$ & $51.42$ & $51.60$ & ${\bf 56.16}$\\
                        & $\alpha=0.5$ & $54.32$ & $54.99$ & $56.09$ & $53.4$ & $55.16$ & ${\bf 57.89}$\\
                        & $\alpha=1.0$ & $55.69$ & $55.58$ & $58.13$ & $56.34$ & $56.82$ & ${\bf 61.45}$\\
\hline
\multirow{3}{*}{CIFAR-100} & $\alpha=0.1$ & $29.95$ & $29.76$ & $31.77$ & $29.01$ & $29.86$   & ${\bf 33.82}$\\
                       & $\alpha=0.5$     & $31.03$ & $32.97$ & $34.06$ & $30.31$ & $33.96$ & ${\bf 36.85}$\\
                       & $\alpha=1.0$      & $33.05$ & $33.74$ & $35.81$ & $31.88$ & $33.08$ & ${\bf 37.26}$\\
\hline
\multirow{1}{*}{FEMNIST} & - & $82.32$ & $82.61$ & $81.88$ & $82.86$  & $81.68$  & ${\bf 83.94}$\\
\hline
\end{tabular}
\caption{Test accuracy on various partitions of MNIST, CIFAR10, CIFAR100, and FEMNIST}
\label{tab:acc}
	\vspace{-0.1in}
\end{table*}

\subsection{Convergence Analysis}
We provide insights into the convergence analysis for FedCat
based on the following assumptions about 
the loss functions of  local devices.
\newtheorem{assumption}{Assumption}[section]
\begin{assumption}\label{asm1}
For $i = 1, 2, \cdots, N$, $ f_i $ is L-smooth satisfying $|| \nabla f_i(x) - \nabla f_i(y) || \leq l ||x - y||$.
\end{assumption}
\begin{assumption}\label{asm2}
For $i = 1, 2, \cdots, N$, $ f_i $ is $\mu$-convex satisfying $|| \nabla f_i(x) - \nabla f_i(y) || \geq \mu ||x - y||$, where $\mu \geq 0$.
\end{assumption}
\begin{assumption}\label{asm3}
The variance of stochastic gradients is upper bounded by  $\beta^2$ and the expectation of squared norm of stochastic gradients is upper bounded by  $G^2$, i.e.,  $\mathbb{E}||\nabla f_k (w;\xi) - \nabla f_k (w) ||^2 \leq \beta^2$, $\mathbb{E}||\nabla f_k (w;\xi) ||^2 \leq G^2$, where $\xi$ is a  data batch  of the $k^{th}$ device in the $t^{th}$ iteration.
\end{assumption}

In the case where all the devices participate in every round of local training, inspired by the work in \cite{convergence}, we can derive
  Theorem \ref{thm1}: 
\newtheorem{thm}{\bf Theorem}[section]
\begin{thm}\label{thm1}
Let Assumption \ref{asm1}, Assumption \ref{asm2}, Assumption \ref{asm3} hold. After the  aggregation in each round, 
we have
\begin{footnotesize}
\vspace{-0.1in}
\begin{equation}
\begin{split}
    \mathbb{E}||F(\overline{w}_t)|| - F^\star \leq \frac{L}{\mu(\gamma + t - 1)} (\frac{2 B}{\mu} + \frac{\mu \gamma}{2}\mathbb{E}||\overline{w}_1 - w^\star||^2)
     \nonumber
\end{split},
\end{equation}
\end{footnotesize}
where
\begin{footnotesize}
$
    B = \frac{1}{N} \beta^2 + 6 L \Gamma + 8(K E - 1)^2 G^2.
     \nonumber
$
\end{footnotesize}
\end{thm} 


Theorem \ref{thm1} indicates that the difference between the current loss $F(\overline{w}_t)$ and the optimal loss $F^\star$ is inversely related to $t$. From Theorem \ref{thm1}, we can find that the convergence rate of FedCat is consistent with that of FedAvg, which is analyzed in \cite{convergence}. 
Please refer to Appendix \ref{sec:appendix} for more details of the proof.

\section{Experimental Results}\label{sec:experiment}

To evaluate the performance of our  FedCat method, we compared both the model accuracy and convergence of FedCat with five baselines using
four well-known benchmarks.
Moreover, we designed ablation studies to demonstrate the effectiveness of our proposed device concatenation-based training and  device selection strategies.

\begin{figure*}[h]
	\centering
	\subfigure[CIFAR-10 with $\alpha=0.1$]{
		\centering
		\includegraphics[width=0.25\textwidth]{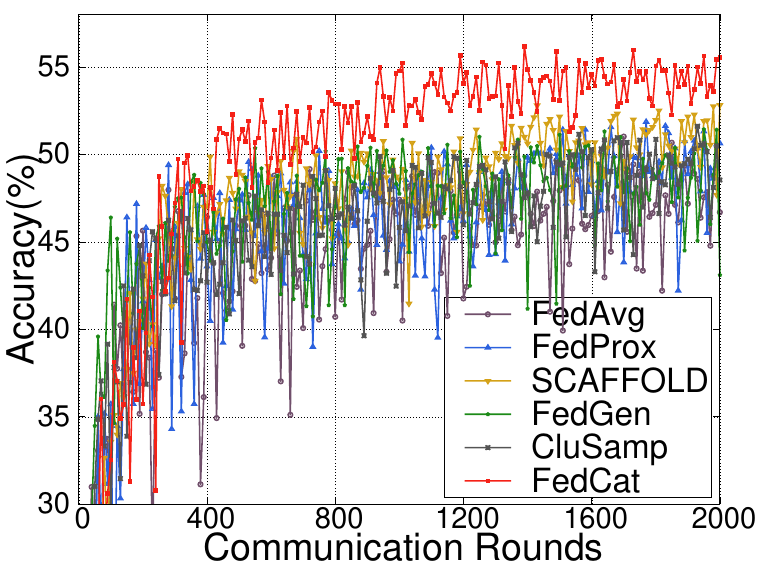}
		\label{fig:cifar-10-0.1}
	}\hspace{-0.15in}
	\subfigure[CIFAR-10  with $\alpha=0.5$]{
		\centering
		\includegraphics[width=0.25\textwidth]{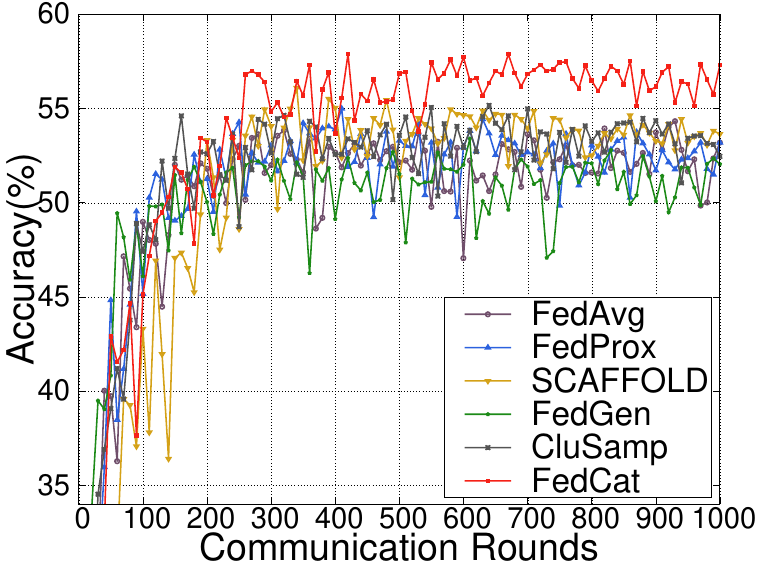}
		\label{fig:cifar-10-0.5}
	}\hspace{-0.15in}
	\subfigure[CIFAR-10 with $\alpha=1.0$]{
		\centering
		\includegraphics[width=0.25\textwidth]{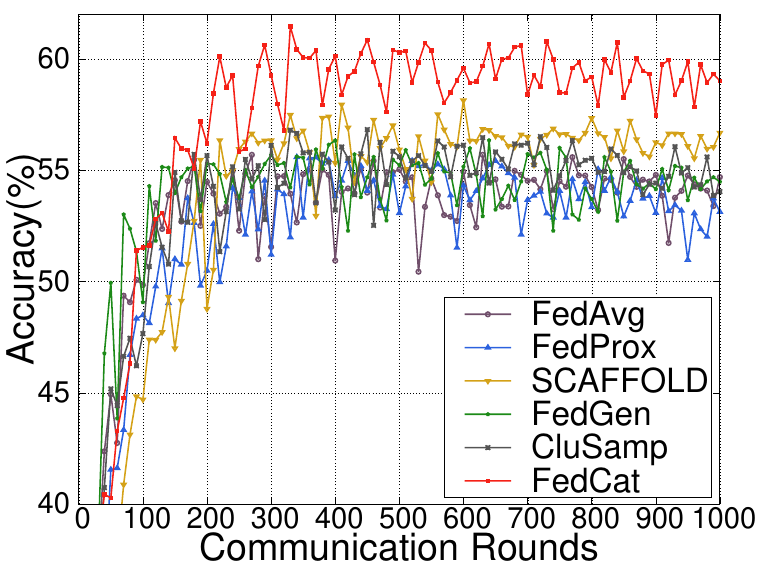}
		\label{fig:cifar-10-1.0}
	}\hspace{-0.15in}
	\subfigure[FEMNIST]{
		\centering
		\includegraphics[width=0.25\textwidth]{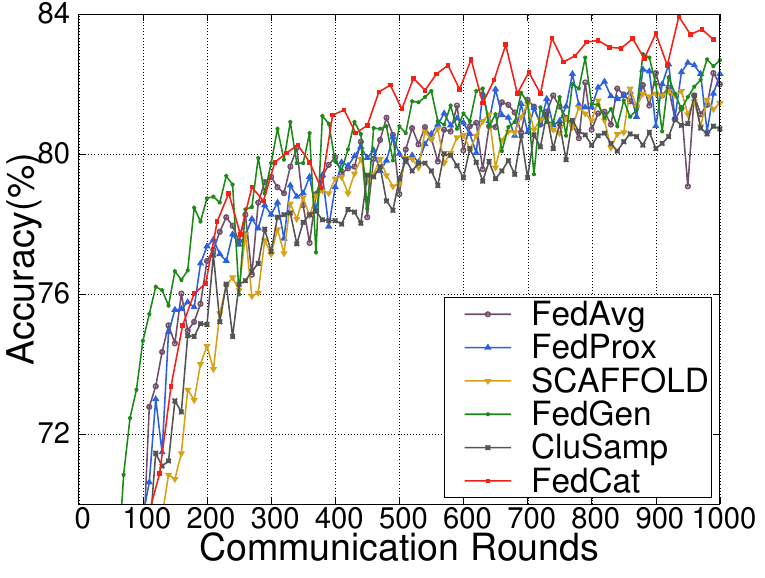}
		\label{fig:femnist}
	}

	\vspace{-0.1in}
	\caption{Comparison results   between of our FedCat approach and five baseline methods}
	\label{fig:accuacy}
		\vspace{-0.1in}
\end{figure*}

\subsection{Experimental Settings}

We implemented our approach on top of a cloud server and a large set of  local devices.
Since  it is difficult for all the devices to participate in the training process simultaneously in practice, by default we  assumed that there are only 10\% of the overall    local devices 
selected to participate in each round of model training.
We used the SGD optimizer with a learning rate 0.01, where the SGD momentum is 0.9 for all the methods.
For each device, we set the batch size of its local training  to 50,  and performed 
five epochs for each local training. 
We tested the accuracy of the global model in every ten rounds, where each round involves a pair of model upload and dispatch
operations. We performed experiments on a workstation with Intel i9-10900k CPU, 32GB memory, NVIDIA GeForce RTX 3080 GPU, and Ubuntu operating system (version 18.04).

{\bf Dataset and Model Settings}: Our experiments were conducted on four well-known datasets, i.e., 
MNIST, CIFAR-10, CIFAR-100~\cite{data}, and FEMNIST~\cite{leaf}.
For the first three datasets, we assumed that there are 100 devices got involved in FL. 
For  FEMNIST,  we considered a non-IID scenario with 180  devices, where each  consists of more than 100 local samples\footnote{Using the following 
command in the LEAF benchmark: ./preprocess.sh -s niid –sf 0.05 -k 100 -t sample. }.
For MNIST and CIFAR-10, we adopted
the same  CNN models as the ones used  in \cite{communication}.
We use the  CNN model of MNIST for FEMNIST with slight modification. 
Since the number of categories of  FEMNIST is 62, we changed the 
output size of MNIST CNN model to 62 for FEMNIST.
For  CIFAR-100, we modified the output size of the CIFAR-10 CNN model to 20,
since there are 20  categories of  coarse-grained labels in  CIFAR-100.

{\bf Data Heterogeneity Settings}:
In the experiment, we adopted
the Dirichlet distribution~\cite{measuring} to control the heterogeneity of device data
for MNIST, CIFAR-10, and CIFAR-100.
Here, we use the notation $Dir(\alpha)$ to denote different Dirichlet distributions
controlled by $\alpha$, where  a smaller value of $\alpha$ indicates higher data heterogeneity. 
For each one of datasets MNIST, CIFAR-10, and CIFAR-100, we investigated
three cases following  different Dirichlet distributions with 
$\alpha=0.1$, $0.5$, and $1.0$, respectively. 
Note  that, different from the other three datasets,  the raw data of FEMNIST is naturally non-IID distributed considering various kinds of imbalances including class imbalance, data imbalance, and data heterogeneity.

{\bf Baseline Methods}: To fairly evaluate the performance of our approach, we compared the model accuracy of FedCat with five baseline methods, i.e.,  FedAvg~\cite{communication}, FedProx~\cite{fedprox}, SCAFFOLD~\cite{scaffold}, FedGen~\cite{datafree}, and CluSamp~\cite{clustered}. FedAvg is the most classical FL method and the other four methods are the state-of-the-art
representatives of the three kinds of FL  methods introduced in Section~\ref{sec:relatedwork}. 
FedProx and SCAFFOLD are global control variable-based methods.
FedGen is a KD-based approach, and CluSamp is a device grouping-based method.
Note that,  FedProx uses a hyperparameter $\mu$ to control the weight of its proximal term. 
When using FedProx, we only considered four values for $\mu$ defined in the set \{0.001,0.01,0.1,1\}. 
We explored the best values of $\mu$ for MNIST, CIFAR-10, CIFAR-100, and FEMNIST, which 
 0.1, 0.01, 0.001, and 0.1, respectively.
For FedGen, we used the same   global generator training setting  as \cite{datafree} for the server.
For CluSamp, we clustered devices based on the similarity of model gradients in the same way as~\cite{clustered}.

\subsection{Performance Comparison}
To evaluate the performance of our approach, we compared our approach with five baselines on four datasets.
Due to the space limitation, here we only present the comparison results for CIFAR-10. All the supplementary 
experimental results are included
in  Appendix~\ref{sec:appendix_exp}.

{\bf Comparison of Accuracy:}
Table~\ref{tab:acc} presents  the performance
comparison results between our  FedCat and the five  baseline methods in terms of test accuracy. 
Note that the second column shows the values of the hyperparameter $\alpha$ that control the data heterogeneity based on the  Dirichlet distribution. 
From Table~\ref{tab:acc}, we can observe that compared with the  five baseline methods, our approach can archive the highest accuracy on all  the scenarios with different data heterogeneity settings. 
As an example of CIFAR-10, when $\alpha$ equals 0.1, our  FedCat outperforms 
 FedAvg, FedProx, SCAFFOLD, FedGen, and CluSamp by  5.15\%, 4.31\%, 3.35\%, 4.74\%, and 4.56\%, respectively.
Note that, our approach can achieve better
improvements on the datasets 
CIFAR-10 and CIFAR-100 than the datasets MINST and FEMNIST. 
This is because the datasets MINST and FEMNIST are  much simpler 
than the datasets CIFAR-10 and CIFAR-100, where all the FL methods in the table can achieve almost the best 
test accuracy.

Figure \ref{fig:accuacy} presents the model accuracy trends of all the investigated FL methods considering different 
data heterogeneity settings for datasets CIFAR-10 and FEMNIST. Note that here 
one communication round indicates one interaction between the server and devices including both the gradient upload and 
model dispatch operations. From this figure, we can find that FedCat cannot only achieve the highest 
accuracy, but also converge more quickly than the other methods to  their highest accuracy.

{\bf Analysis of Communication Overhead:}
Similar to FedAvg,  FedProx, and CluSamp, in one round of FL, the communication overhead of FedCat only involves the network traffic generated by  
both model dispatching and gradient uploading operations. 
Due to the usage of extra global control variables, SCAFFOLD needs twice as much network traffic as FedCat.
For FedGen, the communication overhead is more than FedAvg, since it needs to dispatch an additional built-in generator with a non-negligible size compared to the model itself. 
In a nutshell,  the communication overhead required by FedCat 
is the least among all the six FL methods.

\subsection{Ablation Study}
We conducted  ablation studies to demonstrate the effectiveness of both of our proposed device concatenation-based model training and grouping- and count-based device selection.
We use the notations ``{\bf FedCat w/ GC}'' to denote the  FedCat variant implemented  with only grouping- and count-based device selection strategy 
and ``{\bf FedCat w/ DC}'' to indicate the FedCat variant implemented  with only device concatenation-based model training.
For {\bf FedCat w/ GC}, we aggregate models in every FL round.
For {\bf FedCat w/ DC}, we randomly select devices without grouping.
Note that,
we consider the FedAvg in the ablation studies, since 
FedAvg is  a specific FedCat variant 
 without considering    
device concatenation-based model training or grouping- and count-based device selection.

\begin{figure}[h]
	\begin{center}
		\includegraphics[width=2.35 in]{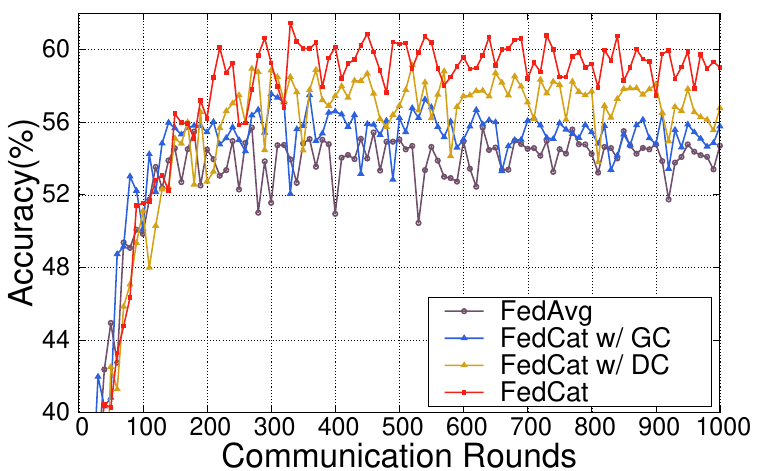}
			\vspace{-0.1in}
		\caption{Ablation studies of model accuracy} \label{fig:cur}
	\end{center}
		\vspace{-0.15in}
\end{figure}

Due to the space limitation,  Figure~\ref{fig:cur} only shows the ablation study results
on CIFAR-10 with $\alpha=1.0$. 
 We can observe that FedCat can achieve the highest model accuracy, 
while both {\bf FedCat w/ GC} and {\bf FedCat w/ DC} have better model accuracy than FedAvg.
Please refer to Appendix \ref{sec:appendix_exp} for the ablation study results of the other  datasets.

\section{Conclusion}\label{sec:conclusions}

Although Federated Learning (FL) approaches
are promising to enable collaborative learning for 
a large set  of involved devices without causing 
data privacy leakage, they still
suffer from the problem of model inaccuracy in various 
non-IID scenarios. This is because the local models of 
 traditional FL methods   
are trained on individual devices with insufficient and skewed 
training data, which  can easily lead to weight divergence during the training of  
local models. To address this problem, we proposed a novel FL method named FedCat, which 
allows local models to traverse through a  series of selected 
devices during the local training by using our proposed count-based device selection strategy and 
device concatenation-based local training  method.
Due to the large size of 
training data contained by logically concatenated devices,   FedCat can effectively 
mitigate the problem of model weight divergence during the local training process, thus 
improving the overall FL performance for 
non-IID scenarios. 
Comprehensive experimental results on four well-known benchmarks show the effectiveness of our approach.

\bibliographystyle{named}
\bibliography{ijcai22}

\begin{thebibliography}{}

\bibitem[\protect\citeauthoryear{Bellemare \bgroup \em et al.\egroup
  }{2016}]{count}
M.~Bellemare, S.~Srinivasan, G.~Ostrovski, T.~Schaul, D.~Saxton, and R.~Munos.
\newblock Unifying count-based exploration and intrinsic motivation.
\newblock In {\em Advances in Neural Information Processing Systems (NeurIPS)},
  pages 1471--1479, 2016.

\bibitem[\protect\citeauthoryear{Caldas \bgroup \em et al.\egroup
  }{2018}]{leaf}
S.~Caldas, P.~Wu, T.~Li, J.~Konecny, H.~McMahan, V.~Smith, and A.~Talwalkar.
\newblock {LEAF:} {A} benchmark for federated settings.
\newblock {\em CoRR}, abs/1812.01097, 2018.

\bibitem[\protect\citeauthoryear{Chen \bgroup \em et al.\egroup
  }{2020}]{fedcluster}
C.~Chen, Z.~Chen, Y.~Zhou, and B.~Kailkhura.
\newblock Fedcluster: Boosting the convergence of federated learning via
  cluster-cycling.
\newblock In {\em Proceedings of International Conference on Big Data
  (BigData)}, pages 5017--5026, 2020.

\bibitem[\protect\citeauthoryear{et al.}{2021}]{advance}
P.~Kairouz et~al.
\newblock Advances and open problems in federated learning.
\newblock {\em Found. Trends Mach. Learn.}, 14(1-2):1--210, 2021.

\bibitem[\protect\citeauthoryear{Fraboni \bgroup \em et al.\egroup
  }{2021}]{clustered}
Y.~Fraboni, R.~Vidal, L.~Kameni, and M.~Lorenzi.
\newblock Clustered sampling: low-variance and improved representativity for
  clients selection in federated learning.
\newblock In {\em Proceedings of International Conference on Machine Learning
  (ICML)}, volume 139, pages 3407--3416, 2021.

\bibitem[\protect\citeauthoryear{Hsu \bgroup \em et al.\egroup
  }{2019}]{measuring}
T.~Hsu, H.~Qi, and M.~Brown.
\newblock Measuring the effects of non-identical data distribution for
  federated visual classification.
\newblock {\em CoRR}, abs/1909.06335, 2019.

\bibitem[\protect\citeauthoryear{Hu \bgroup \em et al.\egroup }{2021}]{spars}
R.~Hu, Y.~Gong, and Y.~Guo.
\newblock Federated learning with sparsification-amplified privacy and adaptive
  optimization.
\newblock In {\em Proceedings of International Joint Conference on Artificial
  Intelligence (IJCAI)}, pages 1463--1469, 2021.

\bibitem[\protect\citeauthoryear{Huang \bgroup \em et al.\egroup
  }{2021a}]{mimics}
H.~Huang, F.~Shang, Y.~Liu, and H.~Liu.
\newblock Behavior mimics distribution: combining individual and group
  behaviors for federated learning.
\newblock In {\em Proceedings of International Joint Conference on Artificial
  Intelligence (IJCAI)}, pages 2556--2562, 2021.

\bibitem[\protect\citeauthoryear{Huang \bgroup \em et al.\egroup
  }{2021b}]{crosssilo}
Y.~Huang, L.~Chu, Z.~Zhou, L.~Wang, J.~Liu, J.~Pei, and Y.~Zhang.
\newblock Personalized cross-silo federated learning on non-iid data.
\newblock In {\em Proceedings of Conference on Artificial Intelligence (AAAI)},
  pages 7865--7873, 2021.

\bibitem[\protect\citeauthoryear{Karimireddy \bgroup \em et al.\egroup
  }{2019}]{scaffold}
S.~Karimireddy, S.~Kale, M.~Mohri, S.~Reddi, S.~Stich, and A.~Suresh.
\newblock {SCAFFOLD:} stochastic controlled averaging for on-device federated
  learning.
\newblock {\em CoRR}, abs/1910.06378, 2019.

\bibitem[\protect\citeauthoryear{Li and Wang}{2019}]{fedmd}
D.~Li and J.~Wang.
\newblock Fedmd: Heterogenous federated learning via model distillation.
\newblock {\em CoRR}, abs/1910.03581, 2019.

\bibitem[\protect\citeauthoryear{Li \bgroup \em et al.\egroup
  }{2020a}]{fedprox}
T.~Li, A.~Sahu, M.~Zaheer, M.~Sanjabi, A.~Talwalkar, and V.~Smith.
\newblock Federated optimization in heterogeneous networks.
\newblock In {\em Proceedings of Machine Learning and Systems (MLSys)}, 2020.

\bibitem[\protect\citeauthoryear{Li \bgroup \em et al.\egroup
  }{2020b}]{convergence}
X.~Li, K.~Huang, W.~Yang, S.~Wang, and Z.~Zhang.
\newblock On the convergence of fedavg on non-iid data.
\newblock In {\em Proceedings of International Conference on Learning
  Representations (ICLR)}, 2020.

\bibitem[\protect\citeauthoryear{Li \bgroup \em et al.\egroup }{2021}]{oneshot}
Q.~Li, B.~He, and D.~Song.
\newblock Practical one-shot federated learning for cross-silo setting.
\newblock In {\em Proceedings of International Joint Conference on Artificial
  Intelligence (IJCAI)}, pages 1484--1490, 2021.

\bibitem[\protect\citeauthoryear{Lin \bgroup \em et al.\egroup
  }{2020}]{ensembledist}
T.~Lin, L.~Kong, S.~Stich, and M.~Jaggi.
\newblock Ensemble distillation for robust model fusion in federated learning.
\newblock In {\em Proceedings of Annual Conference on Neural Information
  Processing Systems 2020 (NeurIPS)}, 2020.

\bibitem[\protect\citeauthoryear{McMahan \bgroup \em et al.\egroup
  }{2017}]{communication}
B.~McMahan, E.~Moore, D.~Ramage, S.~Hampson, and B.~Arcas.
\newblock Communication-efficient learning of deep networks from decentralized
  data.
\newblock In {\em Proceedings of International Conference on Artificial
  Intelligence and Statistics (AISTATS)}, volume~54, pages 1273--1282, 2017.

\bibitem[\protect\citeauthoryear{Stich}{2019}]{sgd}
Sebastian~U. Stich.
\newblock Local {SGD} converges fast and communicates little.
\newblock In {\em Proceedings of International Conference on Learning
  Representations (ICLR)}, 2019.

\bibitem[\protect\citeauthoryear{Sun \bgroup \em et al.\egroup
  }{2021}]{practical}
L.~Sun, J.~Qian, and X.~Chen.
\newblock {LDP-FL:} practical private aggregation in federated learning with
  local differential privacy.
\newblock In {\em Proceedings of International Joint Conference on Artificial
  Intelligence (IJCAI)}, pages 1571--1578, 2021.

\bibitem[\protect\citeauthoryear{TorchvisionData}{2019}]{data}
TorchvisionData.
\newblock Dataset of mnist, fashion-mnist, cifar-10 and cifar-100.
\newblock Website, 2019.
\newblock \url{https://pytorch.org/docs/stable/torchvision/datasets.html}.

\bibitem[\protect\citeauthoryear{Wang \bgroup \em et al.\egroup
  }{2020}]{weight}
H.~Wang, Z.~Kaplan, D.~Niu, and B.~Li.
\newblock Optimizing federated learning on non-iid data with reinforcement
  learning.
\newblock In {\em Proceedings of Conference on Computer Communications
  (INFOCOM)}, pages 1698--1707, 2020.

\bibitem[\protect\citeauthoryear{Wang \bgroup \em et al.\egroup
  }{2021a}]{imbalance}
L.~Wang, S.~Xu, X.~Wang, and Q.~Zhu.
\newblock Addressing class imbalance in federated learning.
\newblock In {\em Proceedings of Conference on Artificial Intelligence (AAAI)},
  pages 10165--10173, 2021.

\bibitem[\protect\citeauthoryear{Wang \bgroup \em et al.\egroup }{2021b}]{fair}
Z.~Wang, X.~Fan, J.~Qi, C.~Wen, C.~Wang, and R.~Yu.
\newblock Federated learning with fair averaging.
\newblock In {\em Proceedings of International Joint Conference on Artificial
  Intelligence (IJCAI)}, pages 1615--1623, 2021.

\bibitem[\protect\citeauthoryear{Xie \bgroup \em et al.\egroup
  }{2021}]{multicenter}
M.~Xie, G.~Long, T.~Shen, T.~Zhou, X.~Wang, J.~Jiang, and C.~Zhang.
\newblock Multi-center federated learning.
\newblock {\em CoRR}, abs/2108.08647, 2021.

\bibitem[\protect\citeauthoryear{Yang}{2021}]{hfl}
H.~Yang.
\newblock {H-FL:} {A} hierarchical communication efficient and
  privacy-protected architecture for federated learning.
\newblock In {\em Proceedings of International Joint Conference on Artificial
  Intelligence (IJCAI)}, pages 479--485, 2021.

\bibitem[\protect\citeauthoryear{Zhu \bgroup \em et al.\egroup
  }{2021}]{datafree}
Z.~Zhu, J.~Hong, and J.~Zhou.
\newblock Data-free knowledge distillation for heterogeneous federated
  learning.
\newblock In {\em Proceedings of International Conference on Machine Learning,
  {ICML}}, volume 139, pages 12878--12889, 2021.

\end{thebibliography}

\newpage

\appendix
\section{Proof of Convergence Analysis}\label{sec:appendix}

In our scenario, all devices have the same weight. $t$ exhibits the $t^{th}$ SGD iteration on the local device. $v$ is the intermediate variable that represents the result of SGD update after exactly one iteration.
The update of FedCat is as follows:
\begin{equation}
v_{t+1}^k = w_t^k - \eta_t \nabla f_k (w_t^k, \xi_t^k),  \nonumber
\end{equation}
where $w_{t}^k$ represents the model of the $k^{th}$ device in the $t^{th}$ iteration. 
\begin{equation}
w_{t+1}^k=\left\{
\begin{array}{rl}
v_{t+1}^k, &if \quad E \nmid t + 1 \\
v_{t+1}^i (i \neq k),&if \quad E \mid t + 1, K E \nmid t + 1\\
\frac{1}{N} \sum_{i=1}^{N} v_{t+1}^i, &if \quad K E \mid t + 1\\
\end{array}
\right.  \nonumber 
\end{equation}
Similar to \cite{sgd}, we define two variables $\overline{v}_t$ and $\overline{w}_t$:
\begin{equation}
\begin{split}
    \overline{v}_t = \frac{1}{N} \sum_{k=1}^{N} v_t^k, 
    \overline{w}_t = \frac{1}{N} \sum_{k=1}^{N} w_t^k.
\end{split} \nonumber
\end{equation}
Here we explain why $\overline{w}_t = \overline{v}_t$. We consider  the following  three cases:
\begin{itemize}
\item If $E \nmid t + 1$, notice that $w_{t+1}^k = v_{t+1}^k$, then $\overline{w}_t = \overline{v}_t$.
\item If $E \mid t + 1$ and $K E \nmid t + 1$, $\overline{w}_{t+1} = \frac{1}{N} \sum_{k=1}^N w_{t+1}^k = \frac{1}{N} \sum_{k=1}^N v_{t+1}^k =\overline{v}_{t+1}$.
\item  If $K E \mid t + 1$, $w_{t+1}^1 = w_{t+1}^2  = \cdots = w_{t+1}^N = \frac{1}{N} \sum_{k=1}^N w_{t+1}^k = \overline{w}_{t+1} = \frac{1}{N} \sum_{k=1}^N v_{t+1}^k = \overline{v}_{t+1}$.
\end{itemize}

Inspired by \cite{convergence}, we make the following definition:
$g_t = \sum_{k=1}^{N} p_k  \nabla f_k (w_t^k; \xi_t^k)$.
$\overline{g}_t = \sum_{k=1}^{N} p_k  \nabla f_k (w_t^k)$.
$\mathbb{E} [g_t] = \overline{g}_t$.
$\overline{v}_{t+1} = \overline{w}_t - \eta_t g_t$.

\newtheorem{lemma}{Lemma}[section]
\begin{lemma} \label{lemma1}
According to Lemma 1 in \cite{convergence},
\begin{equation}
\begin{split}
    \mathbb{E}||\overline{v}_{t+1} - w^\star|| &\leq (1 - \eta_t\mu)\mathbb{E}||\overline{w}_t - w^\star||^2 + \eta_t^2\mathbb{E}||g_t - \overline{g}_t||^2\\
    &+ 6 L \eta_t^2\Gamma + \frac{2}{N} \mathbb{E} \sum_{k = 1}^{N}||\overline{w}_t - w_t^k||^2
\end{split}
\nonumber
\end{equation}
\end{lemma}
\vspace{-0.2in}
\begin{lemma} \label{lemma2}
The variance of $g_t$ is upper bounded, where all devices have the same aggregation weight $\frac{1}{N}$:
\begin{equation}
\begin{split}
\mathbb{E} || g_t - \overline{g}_t ||^2 \leq \frac{1}{N^2} \sum_{k=1}^{N} \beta^2
\end{split}
\nonumber
\end{equation}
\end{lemma}
\begin{proof}
For the independent random variables, the variance of their sum is equivalent to the sum of their variance. 
\begin{equation}
\begin{split}
\mathbb{E}||\sum_{k=1}^{N}(X_k - \mathbb{E} X_k)||^2 = \sum_{k=1}^{N} \mathbb{E}||X_k - \mathbb{E} X_k||^2
\end{split}
\nonumber 
\end{equation}
Here we use $X_k = \frac{1}{N} \nabla f_k(w_t^k;\xi_t^k)$, $\mathbb{E} X_k = \frac{1}{N} \nabla f_k(w_t^k)$.
\begin{equation}
\begin{split}
\mathbb{E} || g_t - \overline{g}_t ||^2 &= \mathbb{E} || \frac{1}{N} \sum_{k=1}^{N} (\nabla f_k (w_t^k, \xi_t^k) - \nabla f_k (w_t^k)) ||^2\\
&= \sum_{k=1}^{N} \frac{1}{N^2} \mathbb{E} || \nabla f_k (w_t^k, \xi_t^k) - \nabla f_k (w_t^k)) ||^2 \\
&\leq \frac{1}{N^2} \sum_{k=1}^{N} \beta^2\\
\end{split}
\nonumber
\end{equation}
\end{proof}

\begin{lemma} \label{lemma3}
Within our configuration, the aggregation occurs every $K E$ iterations. For arbitrary t, there always exists $t_0 \leq t$ while $t_0$ is the nearest aggregation moment to $t$.  As a result, $t - t_0 \leq K E-1$ holds. Given the constraint on learning rate from \cite{convergence}, we know that $\eta_t \leq \eta_{t_0} \leq 2 \eta_t$. It follows that
\begin{equation}
\begin{split}
\mathbb{E} \sum_{k=1}^{N} \frac{1}{N} ||\overline{w}_t - w _t^k||^2 \leq 4\eta_t^2 (K E - 1)^2 G^2.
\end{split}
\nonumber
\end{equation}
\end{lemma}

\begin{proof}
Let m = $\lfloor \frac{t - t_0}{E} \rfloor$ and the concatenated clients set $A = \{a_1, a_2, ... ,k\}$.
\begin{equation}
\begin{split}
||w _t^k - \overline{w}_{t_0}||^2 = &||\sum_{t=t_0}^{t_0 + E - 1} \eta_t \nabla f_{a_1}(w_t^{a_1};\xi_t^{a_1}) + \\
&\sum_{t=t_0 + E}^{t_0 + 2E - 1} \eta_t \nabla f_{a_2}(w_t^{a_2};\xi_t^{a_2}) + \cdots + \\
&\sum_{t=t_0 + m E}^{t} \eta_t \nabla f_k(w_t^k;\xi_t^k)||^2   \\
& \leq  (t - t_0) \sum_{t = t_0}^{t - 1} \eta_t^2 G^2  \\  
& \leq  (K E - 1) \sum_{t = t_0}^{t - 1} \eta_t^2 G^2  \\
\end{split}
\nonumber
\end{equation}
where the first inequality is accomplished via adopting AM-GM inequality. 
\begin{equation}
\begin{split}
\mathbb{E} \sum_{k=1}^{N} \frac{1}{N} ||\overline{w}_t - w _t^k ||^2 &= \mathbb{E} \sum_{k=1}^{N} \frac{1}{N} || (w_t^k - \overline{w}_{t_0}) - (\overline{w}_t - \overline{w}_{t_0}) ||^2 \\
&\leq \mathbb{E} \sum_{k=1}^{N} \frac{1}{N} || w_t^k - \overline{w}_{t_0}||^2 \\
&\leq \sum_{k=1}^{N} \frac{1}{N} \mathbb{E} \sum_{t = t_0}^{t - 1} (K E - 1) \eta_{t_0}^2 || \nabla f_k (w_t^k, \xi_t^k) ||^2 \\
&\leq \sum_{k=1}^{N} \frac{1}{N} \mathbb{E} \sum_{t = t_0}^{t - 1} (K E - 1) \eta_{t_0}^2 G^2 \\
&\leq 4\eta_t^2 (K E - 1)^2 G^2.   
\end{split}
\nonumber
\end{equation}
\end{proof}
From Lemma \ref{lemma1}, Lemma \ref{lemma2}, and Lemma \ref{lemma3}, we can derive Theorem \ref{thm1}:
\begin{equation}
\begin{split}
    \mathbb{E}||F(\overline{w}_t)|| - F^\star \leq \frac{L}{\mu(\gamma + t - 1)} (\frac{2 B}{\mu} + \frac{\mu \gamma}{2}\mathbb{E}||\overline{w}_1 - w^\star||^2)
\end{split},\nonumber
\end{equation}
where
$
    B = \frac{1}{N} \beta^2 + 6 L \Gamma + 8(K E - 1)^2 G^2.
$

\begin{table*}[h]
\centering
\small
\begin{tabular}{|c|c|c|c|c|c|c|c|}
\hline
\multirow{2}{*}{Dataset} & Heterogeneity & \multicolumn{6}{c|}{Test Accuracy (\%)} \\
\cline{3-8}
 & Settings & FedAvg & FedProx &  SCAFFOLD & FedGen & CluSamp & FedCat (Ours)\\
\hline
\hline
\multirow{3}{*}{MNIST} & $\alpha=0.1$      & $99.04$    & $99.02$     & $98.92$       & $99.18$  & $98.98$  & ${\bf 99.21}$\\
                       & $\alpha=0.5$      & $99.31$    & $99.27$     & $99.28$       & $99.41$ & $99.27$ & ${\bf 99.42}$\\
                       & $\alpha=1.0$      & $99.37$    & $99.30$     & $99.32$       & $99.37$  & $99.33$  & ${\bf 99.4}$\\
\hline
\multirow{3}{*}{CIFAR-10} & $\alpha=0.1$& $51.01$& $51.85$& $52.81$ & $51.42$ & $51.60$ & ${\bf 56.16}$\\
                        & $\alpha=0.5$ & $54.32$ & $54.99$ & $56.09$ & $53.4$ & $55.16$ & ${\bf 57.89}$\\
                        & $\alpha=1.0$ & $55.69$ & $55.58$ & $58.13$ & $56.34$ & $56.82$ & ${\bf 61.45}$\\
\hline
\multirow{3}{*}{CIFAR-100} & $\alpha=0.1$ & $29.95$ & $29.76$ & $31.77$ & $29.01$ & $29.86$   & ${\bf 33.82}$\\
                       & $\alpha=0.5$     & $31.03$ & $32.97$ & $34.06$ & $30.31$ & $33.96$ & ${\bf 36.85}$\\
                       & $\alpha=1.0$      & $33.05$ & $33.74$ & $35.81$ & $31.88$ & $33.08$ & ${\bf 37.26}$\\
\hline
\multirow{1}{*}{FEMNIST} & - & $82.32$ & $82.61$ & $81.88$ & $82.86$  & $81.68$  & ${\bf 83.94}$\\
\hline
\end{tabular}
\caption{Test accuracy on various data distributions of MNIST, CIFAR10, CIFAR100, and FEMNIST}
\label{tab:allacc}
\end{table*}

\section{Supplementary Experimental Results for Accuracy Comparison and Ablation Study}\label{sec:appendix_exp}
\subsubsection{Experimental Results for Accuracy Comparison}
\begin{figure*}[h]
	\centering
	\subfigure[MNIST $\alpha=0.1$]{
		\centering
		\includegraphics[width=0.32\textwidth]{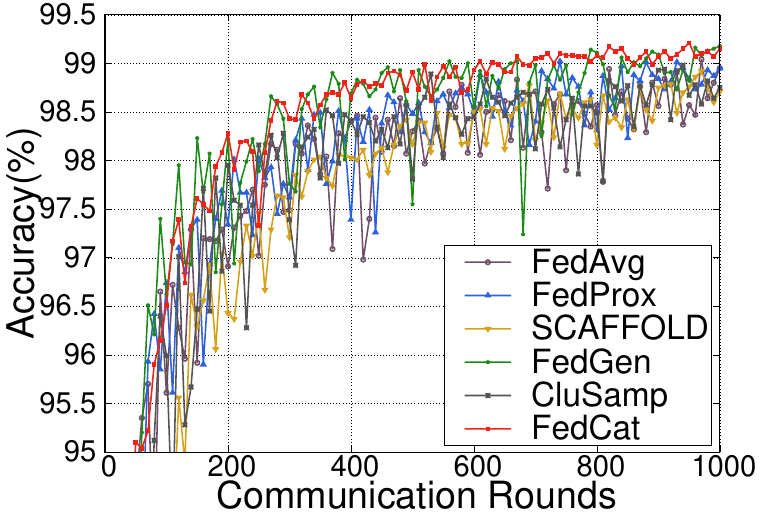}
		\label{fig:MNIST-10-0.1}
	}
	\subfigure[MNIST $\alpha=0.5$]{
		\centering
		\includegraphics[width=0.32\textwidth]{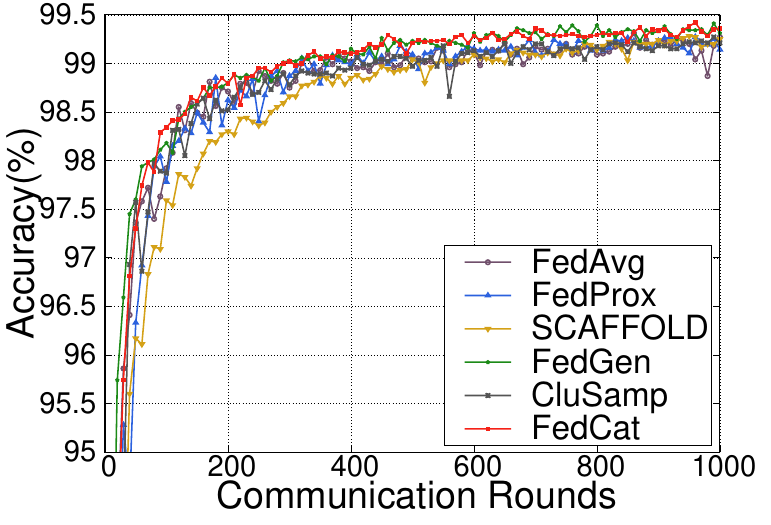}
		\label{fig:MNIST-10-0.5}
	}
	\subfigure[MNIST $\alpha=1.0$]{
		\centering
		\includegraphics[width=0.32\textwidth]{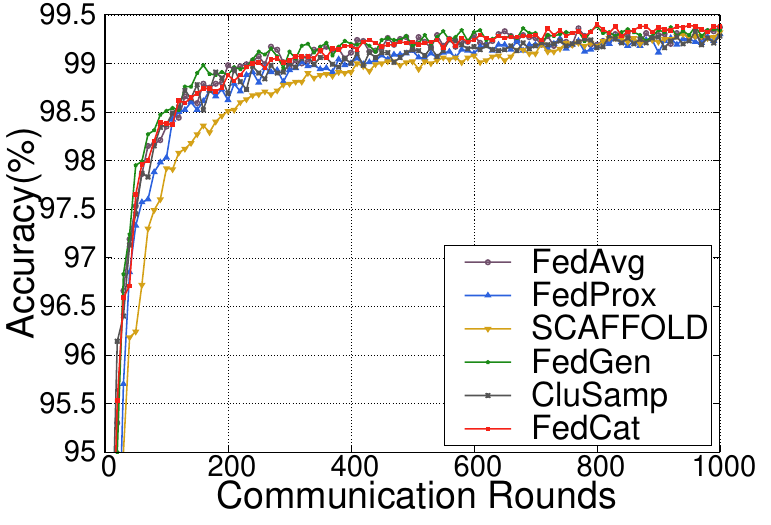}
		\label{fig:MNIST-10-1.0}
	}
	\subfigure[CIFAR-10 $\alpha=0.1$]{
		\centering
		\includegraphics[width=0.32\textwidth]{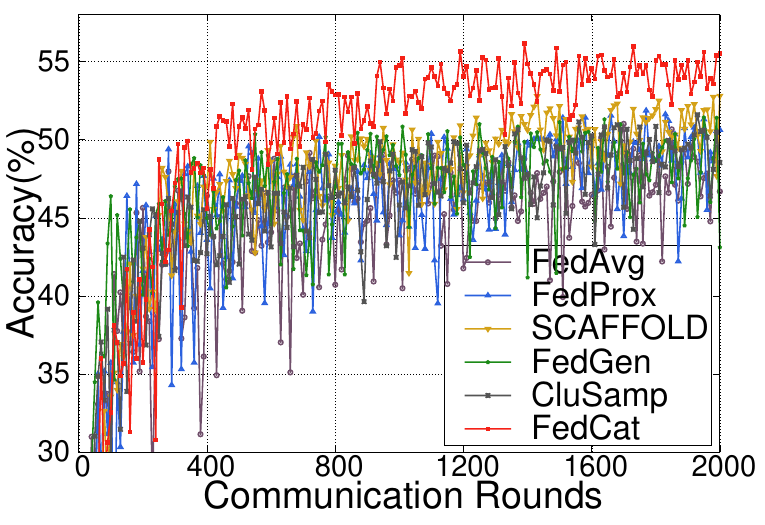}
		\label{fig:cifar-10-0.1-APPENDIX}
	}
	\subfigure[CIFAR-10 $\alpha=0.5$]{
		\centering
		\includegraphics[width=0.32\textwidth]{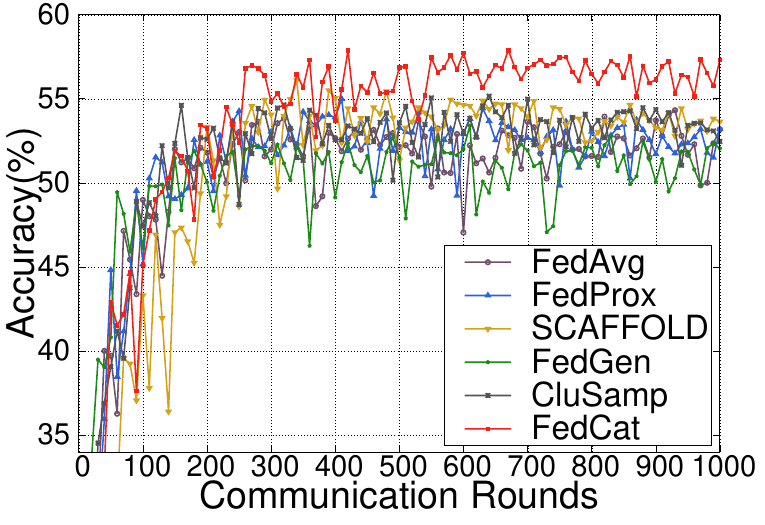}
		\label{fig:cifar-10-0.5-APPENDIX}
	}
	\subfigure[CIFAR-10 $\alpha=1.0$]{
		\centering
		\includegraphics[width=0.32\textwidth]{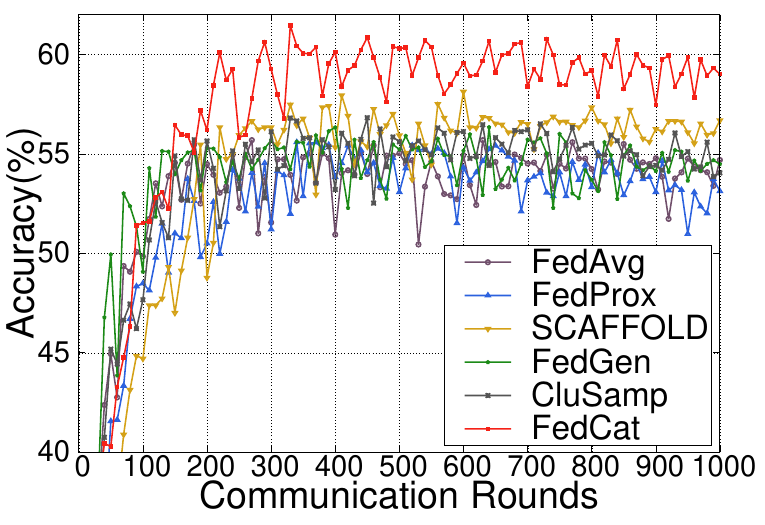}
		\label{fig:cifar-10-1.0-APPENDIX}
	}
	\subfigure[CIFAR-100 $\alpha=0.1$]{
		\centering
		\includegraphics[width=0.32\textwidth]{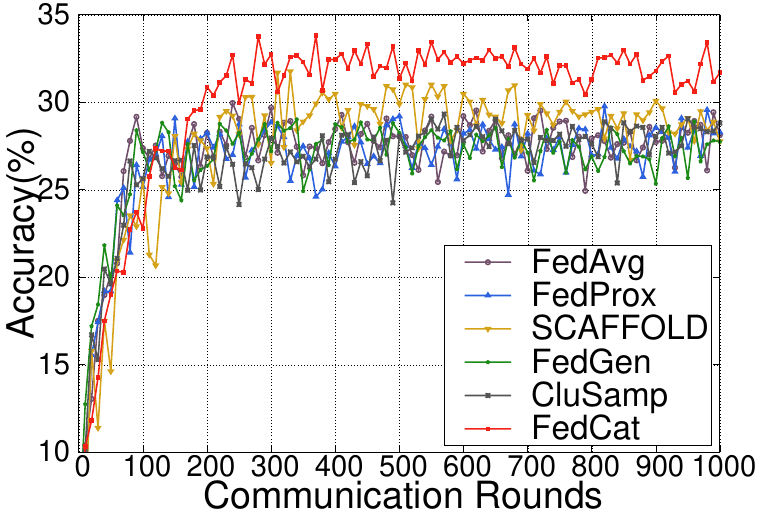}
		\label{fig:cifar-100-0.1-APPENDIX}
	}
	\subfigure[CIFAR-100 $\alpha=0.5$]{
		\centering
		\includegraphics[width=0.32\textwidth]{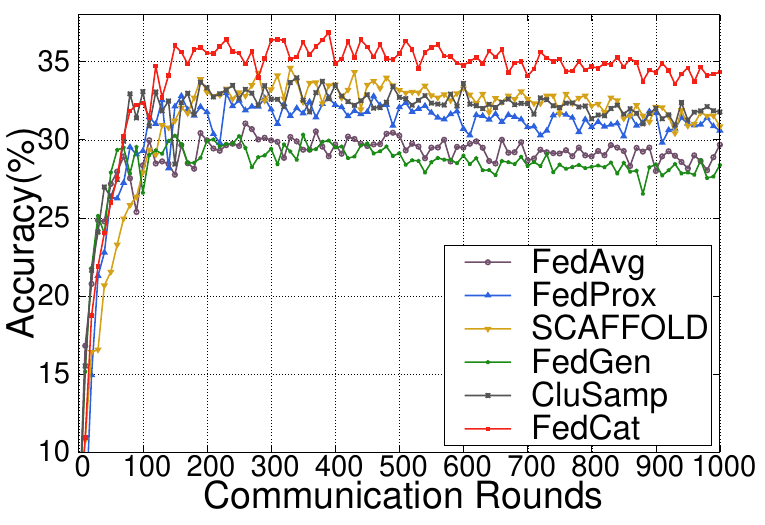}
		\label{fig:cifar-100-0.5-APPENDIX}
	}
	\subfigure[CIFAR-100 $\alpha=1.0$]{
		\centering
		\includegraphics[width=0.32\textwidth]{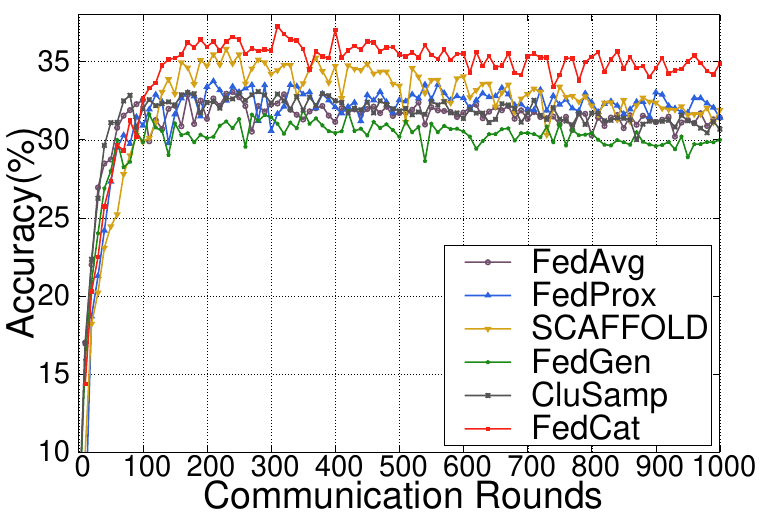}
		\label{fig:cifar-100-1.0-APPENDIX}
	}
	\subfigure[FEMNIST]{
		\centering
		\includegraphics[width=0.32\textwidth]{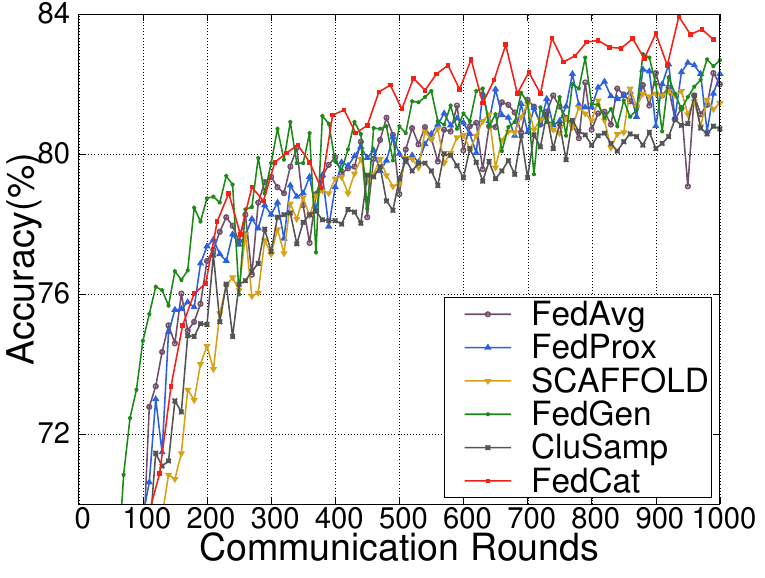}
		\label{fig:femnist-appendix}
	}
\vspace{-0.1in}
	\caption{Model accuracy comparison between our approach and baselines}
\label{fig:acc_appendix}
\end{figure*}

Table \ref{tab:allacc} presents test accuracy for our proposed FedCat and five baselines in 10 scenarios. 
Figure \ref{fig:acc_appendix} presents the trend of accuracy of our approach and baselines.

\subsubsection{Experimental Results of Ablation Studies}
\begin{figure*}[h]
	\centering
	\subfigure[CIFAR-10 $\alpha=0.1$]{
		\centering
		\includegraphics[width=0.32\textwidth]{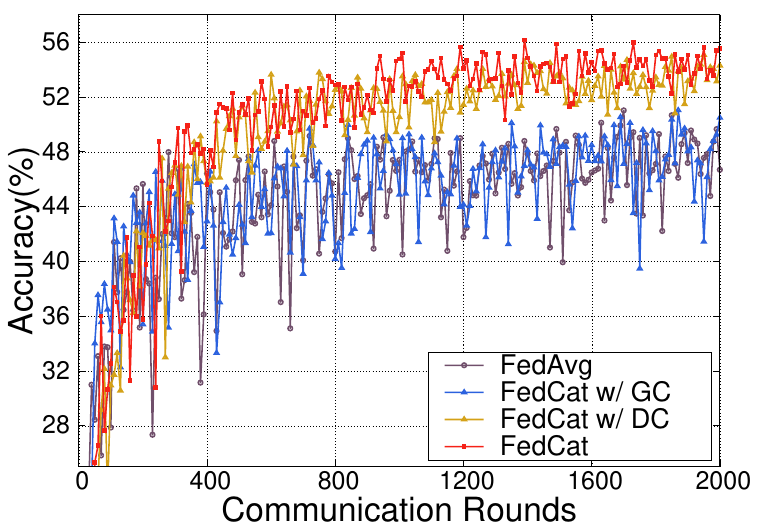}
		\label{fig:cifar-10-0.1-APPENDIX}
	}
	\subfigure[CIFAR-10 $\alpha=0.5$]{
		\centering
		\includegraphics[width=0.32\textwidth]{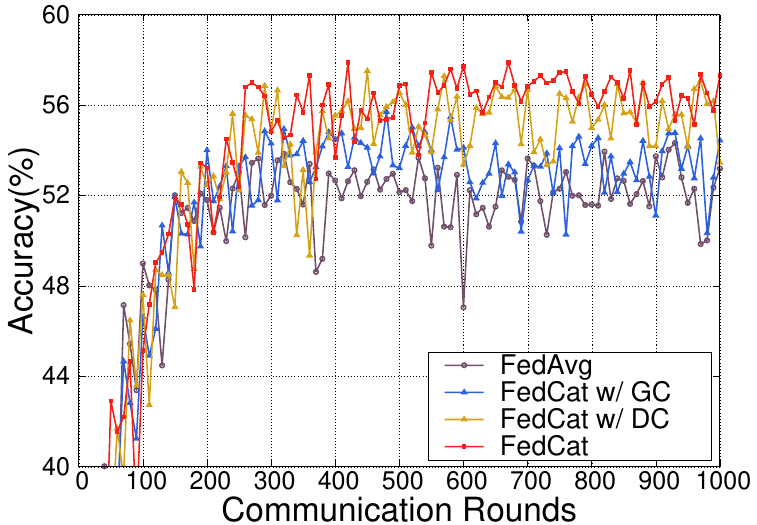}
		\label{fig:cifar-10-0.5-APPENDIX}
	}
	\subfigure[CIFAR-10 $\alpha=1.0$]{
		\centering
		\includegraphics[width=0.32\textwidth]{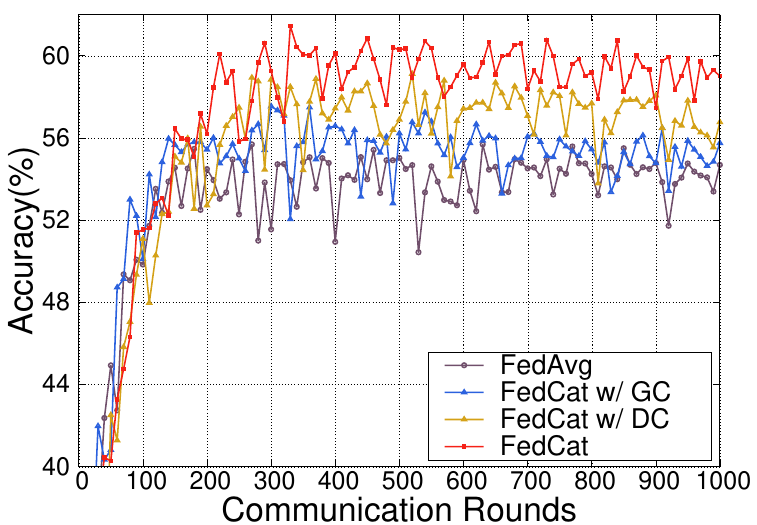}
		\label{fig:cifar-10-1.0-APPENDIX}
	}
	\subfigure[CIFAR-100 $\alpha=0.1$]{
		\centering
		\includegraphics[width=0.32\textwidth]{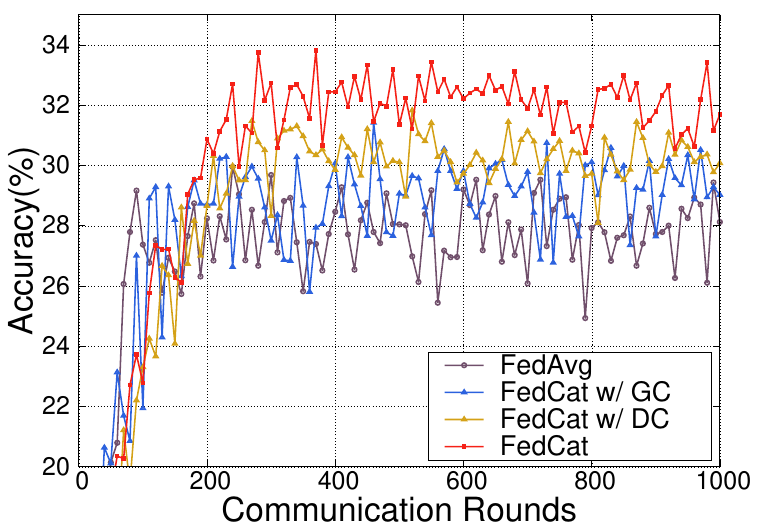}
		\label{fig:cifar-100-0.1-APPENDIX}
	}
	\subfigure[CIFAR-100 $\alpha=0.5$]{
		\centering
		\includegraphics[width=0.32\textwidth]{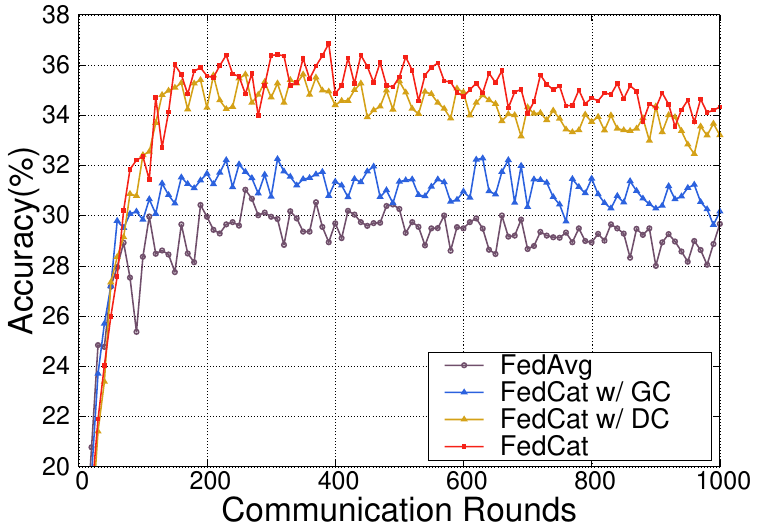}
		\label{fig:cifar-100-0.5-APPENDIX}
	}
	\subfigure[CIFAR-100 $\alpha=1.0$]{
		\centering
		\includegraphics[width=0.32\textwidth]{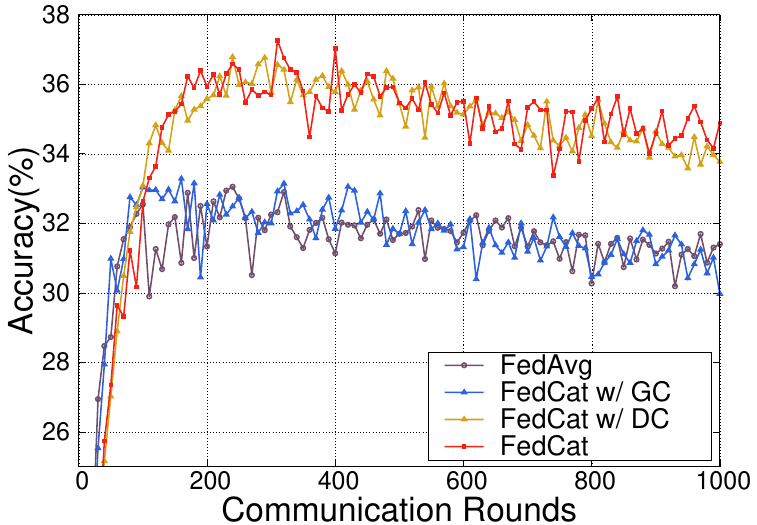}
		\label{fig:cifar-100-1.0-APPENDIX}
	}
\vspace{-0.1in}
	\caption{Ablation studies of model accuracy}
\label{fig:abl_appendix}
\end{figure*}

Table \ref{tab:allabl} presents the test accuracy of our ablation studies.
Figure \ref{fig:abl_appendix} presents the trend of accuracy of our ablation studies on CIFAR-10 and CIFAR-100.

\begin{table}[h]
\centering
\small
\begin{tabular}{|c|c|c|c|c|c|}
\hline
\multirow{3}{*}{Dataset} & \multirow{2}{*}{Heterogeneity} & \multicolumn{4}{c|}{Test Accuracy (\%)} \\
\cline{3-6}
 & \multirow{2}{*}{Settings} & \multirow{2}{*}{FedAvg} & FedCat &  FedCat  & \multirow{2}{*}{FedCat} \\
 & & & w/ GC & w/ DC & \\
\hline
\hline
\multirow{3}{*}{CIFAR-10} & $\alpha=0.1$& $51.01$& $51.02$& $55.06$ & ${\bf 56.16}$\\
                        & $\alpha=0.5$ & $54.32$ & $57.20$ & $57.51$ & ${\bf 57.89}$\\
                        & $\alpha=1.0$ & $55.69$ & $57.54$ & $59.12$ & ${\bf 61.45}$\\
\hline
\multirow{3}{*}{CIFAR-100} & $\alpha=0.1$ & $29.95$ & $31.41$ & $31.82$   & ${\bf 33.82}$\\
                       & $\alpha=0.5$     & $31.03$ & $32.28$ & $35.62$ & ${\bf 36.85}$\\
                       & $\alpha=1.0$      & $33.05$ & $33.28$ & $36.78$ & ${\bf 37.26}$\\
\hline
\end{tabular}
\caption{Test accuracy of ablation studies}
\label{tab:allabl}
\end{table}

\end{document}